\documentclass[num-refs]{wiley-article}
\usepackage{siunitx}
\usepackage{amsmath}
\usepackage{romannum}
\usepackage{mathtools}
\usepackage{adjustbox}
\usepackage{graphicx}
\usepackage{xcolor}
\usepackage{dblfloatfix}
\usepackage{amssymb}
\usepackage{algorithm}
\usepackage{algorithmic}
\usepackage{mathrsfs}

\newcommand{\Cb}{{\mathbf{C}}}
\newcommand{\Db}{{\mathbf D}}
\newcommand{\Eb}{{\mathbf E}}

\newcommand{\Gb}{{\mathbf G}}

\newcommand{\Ib}{{\mathbf I}}

\newcommand{\Pb}{{\mathbf P}}

\newcommand{\Zb}{{\mathbf Z}}

\newcommand{\gb}{{\mathbf g}}
\newcommand{\hb}{{\mathbf h}}

\newcommand{\kb}{{\mathbf k}}

\newcommand{\yb}{{\mathbf y}}
\newcommand{\zb}{{\mathbf z}}

\newcommand{\Phib}{{\boldsymbol {\Phi}}}
\newcommand{\Psib}{{\boldsymbol {\Psi}}}

\newcommand{\Lambdab}{{\boldsymbol {\Lambda}}}

\newcommand{\Rd}{{\mathbb R}}
\newcommand{\Cd}{{\mathbb C}}

\newcommand{\phib}{{\boldsymbol{\phi}}}
\newcommand{\psib}{{\boldsymbol{\psi}}}

\newcommand{\Kc}{{{\mathcal K}}}
\newcommand{\Pc}{{{\mathcal P}}}

\newcommand{\rank}{\textsc{Rank}}
\newcommand{\hank}{\mathscr{H}}

\newcommand{\zerob}{\mathbf{0}}

\newcommand{\blmath}[1]{\mathbf{#1}}

\newcommand{\B}{\blmath{B}}

\newcommand{\E}{\blmath{E}}

\newcommand{\Z}{\blmath{Z}}

\newcommand{\z}{\blmath{z}}

\papertype{Original Article}
\paperfield{Journal Section}

\title{$k$-Space Deep Learning for Reference-free EPI Ghost Correction}
\abbrevs{EPI, echo planar imaging; ALOHA, annihilating low rank Hankel matrix approach
}
\wordcount{approximate 5000 words.}

\author[1]{Juyoung Lee}
\author[1]{Yoseob Han}
\author[2,3]{Jae-Kyun Ryu}
\author[2,3]{Jang-Yeon Park}
\author[1]{Jong Chul Ye}

\affil[1]{Department of Bio and Brain Engineering, Korea Advanced Institute of Science and Technology (KAIST), Daejeon, South Korea}
\affil[2]{Center for Neuroscience Imaging Research, Institute for Basic Science, Suwon, South Korea}
\affil[3]{Department of Biomedical Engineering Sungkyunkwan University, Suwon, South Korea}

\corraddress{Jong Chul Ye, PhD., Department of Bio and Brain Engineering, Korea Advanced Institute of Science and Technology (KAIST), Rm 1107, CMS (E16) Building, 291 Daehak-ro, Yuseong-gu, Daejeon, 34141 South Korea
}
\corremail{jong.ye@kaist.ac.kr}

\fundinginfo{National Research Foundation of Korea, Grant Number: NRF-2016R1A2B3008104}

\runningauthor{Juyoung Lee et al.}

\begin{document}
	
	\maketitle
	
	\begin{abstract}
		\noindent\textbf{Purpose:} Nyquist ghost artifacts in EPI are originated from phase mismatch between the even and odd echoes. However, conventional correction methods using reference scans often produce erroneous results especially in high-field MRI due to the non-linear and time-varying local magnetic field changes. Recently, it was shown that the problem of ghost correction can be reformulated as $k$-space interpolation problem that can be solved using structured low-rank Hankel matrix approaches. Another recent work showed that data driven Hankel matrix decomposition can be reformulated to exhibit similar structures as deep convolutional neural network. By synergistically combining these findings, we propose a $k$-space deep learning approach that immediately corrects the phase mismatch without a reference scan in both accelerated and non-accelerated EPI acquisitions.\\
		\textbf{Theory and Methods:} 
		To take advantage of the even and odd-phase directional redundancy, the $k$-space data is divided into two channels configured with even and odd phase encodings. The redundancies between coils are also exploited by stacking the multi-coil $k$-space data into additional input channels. Then, our $k$-space ghost correction network is trained to learn the interpolation kernel  to estimate the missing virtual $k$-space   data. For the accelerated EPI data, the same neural network is trained to directly estimate the interpolation kernels for missing $k$-space data from both ghost and subsampling.
		\\
		\textbf{Results:} Reconstruction results using 3T and 7T in-vivo data showed that the proposed method outperformed the image quality compared to the existing methods, and the computing time is much faster.\\
		\textbf{Conclusion:} The proposed $k$-space deep learning for EPI ghost correction is highly robust and fast, and can be combined with acceleration, so that it can be used as a promising correction tool for high-field MRI without changing the current acquisition protocol. \\
		
	\end{abstract}

	\section{Introduction}
	
	Echo-planar imaging (EPI) is one of the widely used MR imaging sequences. The single-shot EPI sequence is fast because the whole $k$-space data is acquired through a single RF pulse by alternating the magnetic field direction in the even and odd lines. This short scan  provides high temporal resolution and has therefore been used in many imaging modalities such as functional MRI, diffusion-weighted imaging (DWI),  and so on. However, the rapid change of the magnetic field direction for each line causes the induction of eddy currents in coils and the magnet housing. These eddy currents, in turn, generate local fields that distort $B_o$ and produce a phase mismatch between even and odd rows. Due to this phase disparity, EPI images suffer from  artifacts often called the Nyquist ghost artifact. Since this artifact overlaps with the original image, it often causes difficulties in analyzing the reconstruction results.
	
	There have been many studies to remove ghost artifact. One of the most widely used  methods  to remove ghost artifact is a navigator-based method \cite{ahn1987newphase,bruder1992image,hu1996artifact,reeder1999novel,heid2000method}. In this method, a pre-scanning, often called as reference or navigator scan, is separately acquired before an EPI scan to compensate for the phase mismatch. A simple approach of navigator-based method is to use  $k$-space lines without phase encoding gradient so that the phase difference between even and odd lines can be calculated. Then, the phase correction is performed across all phase encoding lines assuming that the phase offset varies linearly. In another method \cite{heid2000method}, three echoes are obtained, and the phase correction factor is calculated as the difference between the actual second echo and the interpolated echo from the first and third echoes.
	However, those methods suffer from many disadvantages. Aside from the additional acquisition time from the pre-scan at every imaging slice, the linearity assumption for the phase variation is not sufficiently accurate especially for high-field MR acquisition due to the complicated field inhomogeneity variations. Moreover, the current fMRI protocol acquires reference scan only once at the beginning of whole scan time, so it is difficult to capture time-varying local field fluctuations.
	
	To address ghost artifacts from these nonlinear and  time-varying local field changes, navigator-free methods were proposed. Some researches with pulse sequence modification show the reduction of ghost artifact \cite{xiang2007correction,kellman2006phased,Buonocore2001image,chen2004removal,hoge2010robust,yang1996double,poser2013single}. In addition,  navigator-free methods without any pulse sequence modification have been also studied by optimizing image domain quality metrics \cite{Buonocore2001image,zhang2004reference,skare2006afast,peterson2015acquisition}. However, these classic navigator-free methods are generally ineffective compared to the reference-based methods,  or the computational complexity is higher. Recently, some correction methods used phased array coils information to correct ghost artifact \cite{Hoge2016dual,xie2018robust,Ianni2018Ghost}.  Dual-polarity generalized autocalibrating partial parallel acquisition (dual polarity GRAPPA)\cite{Hoge2016dual} uses multiple GRAPPA kernels to correct phase mismatch and fill the missing k-space simultaneously. This method is robust to remove artifact from non-linear phase errors, but it requires an additional EPI calibration scan. Also, phase error correction with sensitivity encoding (PEC-SENSE)\cite{xie2018robust} method can effectively remove ghost artifact without a calibration data. This method reconstructs the negative echo and positive echo images by using SENSE. Although PEC-SENSE preserves the image SNR by  correcting ghost artifact from the down-sampled k-space, it always requires accurate coil sensitivity maps to apply SENSE \cite{xie2018robust}.
	
	Another class of high-performance reference-free EPI ghost correction methods  have been proposed using low rank matrix  completion approaches \cite{Lee2016reference,mani2017multi,lobos2017navigator,lobos2018navigator,lyu2018robust,liu2019pecgrappa}. Specifically, Lee et al \cite{Lee2016reference} used the annihilating filter-based low-rank Hankel matrix approach (ALOHA) \cite{jin2016general,ye2017compressive,lee2016acceleration}, where the key idea is to take advantage of the fact that the concatenated Hankel matrix, which consists of even and odd $k$-space lines, has a low-rank structure due to the multi-channel redundancy. Thus, the phase mismatch correction problem in EPI can be reformulated as a missing $k$-space interpolation problem for even and odd $k$-space data that can be solved using low rank Hankel matrix completion. This method can be easily extended to parallel multi-coil imaging  by simply stacking the Hankel matrices of each coil side-by-side and performing low-rank Hankel matrix completion. Recently, the authors in \cite{lobos2018navigator} further extended the idea by formulating the EPI ghost correction problem as a structured low-rank matrix optimization problem with additional constraints, which incorporates the information from both image and k-space domains and exhibits superior performance compared with DPG or other correction methods. Although the low rank-based ghost correction methods such as ALOHA  \cite{Lee2016reference}, LORAKS  \cite{lobos2018navigator}, etc appears flexible, the most important drawbacks of the methods include 1) high computational cost, and 2) incomplete correction for high field EPI data as will be shown later. 
	
	Therefore, the main goal of the proposed approach is to improve the performance of these low-rank structured matrix based ghost artifact correction methods. In particular, this work aims to replace the computationally expensive rank-based optimization methods  with a simple learning based method using convolutional neural network (CNN) without sacrificing the reconstruction quality. This was inspired by the recent mathematical discovery showing that a CNN is related to a data-driven Hankel matrix decomposition \cite{ye2018deep}. Accordingly, similar to the ALOHA-based ghost correction, we formulate the ghost correction problem as the even and odd {\em virtual} $k$-space interpolation problem so that $k$-space deep neural networks can be designed to estimate the interpolation kernels. The redundancy between the coils from the parallel imaging can also be exploited by stacking the multi-coil $k$-space data into multi-channel inputs. 
	
	The proposed CNN-based ghost correction not only reduces reconstruction time to several order of magnitude, but also significantly improves the reconstruction quality, thanks to the amazing expressivity and generalizability of deep neural networks \cite{ye2019understanding}. Moreover, once the network is trained, the neural network does not need any calibration data and training data at all, which makes the algorithm very flexible and practical. Finally, the proposed method can be easily combined with accelerated EPI data acquisition using sub-sampled $k$-space data. In fact, this flexibility comes from the $k$-space learning, in which the goal is to estimate the $k$-space interpolation kernel. Since both the accelerated acquisition and EPI ghost correction can be formulated as $k$-space interpolation kernel estimation problem, the kernels can be readily estimated by our $k$-space neural network. Our extensive experimentation with 3T and 7T in vivo data show that the proposed $k$-space deep learning approaches outperformed existing approaches and even provided robust and accurate ghost corrections for those frames that could not be corrected by ALOHA in both accelerated and non-accelerated acquisitions.

	\section{Theory}
	\subsection{ALOHA-based Ghost Correction}
	
	To make this paper self-contained, we briefly review the ALOHA-based ghost correction \cite{Lee2016reference}.
	Specifically, in the presence of off-resonances $\Delta f(x,y)$, 
	$k$-space measurement from an EPI sequence can be expressed as \cite{hu1996artifact},\cite{poser2013single}:
	\begin{gather}\label{eq:epi}
	\widehat g(k_x,k_y)=
	\int\int \alpha(x,y)e^{\iota 2\pi \left[ \Delta f(x,y) \left( \left( TE+ (n-N/2)ESP \right) + (-1)^n\left( \frac{k_x}{\gamma G_x} \right) \right) \right]}\times e^{\iota 2\pi (k_x x+k_y y)} dxdy
	\end{gather}
	where  $\iota = \sqrt{-1}$ and $k_x$ and $k_y$ directions represent the read-out and phase-encoding, respectively;
	$n$ denotes the phase encoding index  in the EPI echo train of total length $N$, and
	$\alpha(x,y)$ is the transverse magnetization.
	Here, $TE$ is the echo time, and $ESP$ denotes the echo spacing, which is the time between echoes.

	The key idea of ALOHA-based ghost correction \cite{Lee2016reference} is to split the original equation in  \eqref{eq:epi} into the following two sets of {\em virtual} k-space data:
	\begin{eqnarray}
	\widehat g_{+}(k_x,k_y) 
	&=&\int\int \alpha(x,y)e^{\iota 2\pi \left[ \Delta f(x,y) \left( \left( TE+ (n-N/2)ESP \right) + \left( \frac{k_x}{\gamma G_x} \right) \right) \right]}\times e^{\iota 2\pi (k_x x+k_y y)} dxdy \notag\\
	&=&\int\int   A(x,y) e^{ \iota 2\pi \Delta f(x,y)\frac{k_x}{\gamma G_x} } \cdot e^{\iota 2\pi\left(k_x x+k_y y\right)} dxdy \notag  \ ,
	\end{eqnarray}
	and 
	\begin{eqnarray*}
		\widehat g_{-}(k_x,k_y)
		&=&\int\int \alpha(x,y)e^{\iota 2\pi \left[ \Delta f(x,y) \left( \left( TE+ (n-N/2)ESP \right) - \left( \frac{k_x}{\gamma G_x} \right) \right) \right]} \times e^{\iota 2\pi (k_x x+k_y y)} dxdy \notag\\
		&=&\int\int   A(x,y) e^{-\iota 2\pi \Delta f(x,y)\frac{k_x}{\gamma G_x} } \cdot e^{\iota 2\pi\left(k_x x+k_y y\right)} dxdy
	\end{eqnarray*}
	where $$A(x,y) = \alpha(x,y)e^{\iota 2\pi \Delta f(x,y) (TE+ (n-N/2)ESP )}.$$
	Then,  the even and odd echo signals from the actual EPI measurement can be identified as $1/2$-subsampled k-space data from $\widehat g_{+}(k_x,k_y)$ and $\widehat g_{-}(k_x,k_y)$, respectively.
	The image content for the even and odd virtual data are $A(x,y) e^{+\iota 2\pi \Delta f(x,y)\frac{k_x}{\gamma G_x}}$ and 
	$A(x,y)e^{-\iota 2\pi \Delta f(x,y)\frac{k_x}{\gamma G_x} }$, respectively, which are different from each other. While the difference is the main source of the Nyquist ghost artifacts, they differs only in their phase, so there exists significant redundancy between them.
	Thus, ALOHA-based EPI correction \cite{Lee2016reference} exploits these redundancy to interpolate 
	the missing virtual k-space data.

	More specifically, let  $\kb_m =\left(k_x^{(m)},k_y^{(m)}\right), m=1,\cdots, M$ be a collection of finite number of sampling points of the $k$-space confirming to the Nyquist sampling rate. Then,  the discretized $k$-space data  for the virtual even and odd $k$-space data can be defined as
	\begin{eqnarray*}
		\widehat \gb_+  &=& \begin{bmatrix} \widehat g_+(\kb_1) & \cdots  & \widehat g_+(\kb_M) \end{bmatrix}^T ,\\
		\widehat \gb_-  &=& \begin{bmatrix} \widehat g_-(\kb_1) & \cdots  & \widehat g_-(\kb_M) \end{bmatrix}^T  ,
	\end{eqnarray*}
	where the superscript $^T$ denotes the transpose.
	We also define the difference between the virtual $k$-space data:
	\begin{eqnarray}\label{eq:g_delta}
	\widehat \gb_\Delta = \widehat \gb_+ - \widehat \gb_- 
	\end{eqnarray}
	Then, due to the redundancy between
	$A(x,y) e^{+\iota 2\pi \Delta f(x,y)\frac{k_x}{\gamma G_x}}$ and 
	$A(x,y)e^{-\iota 2\pi \Delta f(x,y)\frac{k_x}{\gamma G_x} }$,  their difference is sparse so that
	the $k$-space Hankel matrix constructed using
	$\widehat{\gb}_\Delta$ is low-ranked \cite{ye2017compressive}. 
	This implies that  there exists a vector $\hb\in \Cd^d$ such that
	\begin{eqnarray}\label{eq:an_mat1}
	\hank_d{\left(\widehat \gb_{\Delta}\right)}\hb  &=& \left( \hank_d\left(\widehat \gb_+\right)- \hank_d\left(\widehat\gb_-\right)\right) \hb \notag\\
	& =& \begin{bmatrix} \hank_d\left(\widehat \gb_+\right) &\hank_d\left(\widehat \gb_-\right)\end{bmatrix}\begin{bmatrix} \hb \\ -\hb \end{bmatrix} =\zerob
	\end{eqnarray}
	where $\hank_d\left(\widehat \gb_+\right) \in \Cd^{M\times d}$ (resp. $\hank_d\left(\widehat \gb_-\right) \in \Cd^{M\times d}$) denotes the Hankel matrix with the matrix pencil size $d$ that are constructed using
	$\widehat\gb_+$ (resp. $\widehat\gb_-$). For more details of the Hankel matrix construction, see  \cite{Lee2016reference}.
	This implies that the concatenated Hankel matrix 
	$\begin{bmatrix} \hank_d\left(\widehat \gb_+\right) &\hank_d\left(\widehat \gb_-\right) \end{bmatrix} \in \Cd^{M\times 2d}$
	is also low-ranked. For the case of parallel imaging,   the concatenated Hankel matrix from $P$ parallel coils 
	\begin{eqnarray}\label{eq:Gb}
	\hank_{d|2P}(\widehat\Gb) :=
	\begin{bmatrix} \hank_d\left(\widehat \gb_+^{(1)}\right) &\hank_d\left(\widehat \gb_-^{(1)}\right)  & \cdots & \hank_d\left(\widehat \gb_+^{(P)}\right)& \hank_d\left(\widehat \gb_-^{(P)}\right)\end{bmatrix}   \in \Cd^{M\times {2dP}}
	\end{eqnarray}
	with 
	$\widehat\Gb:=\begin{bmatrix} \widehat\gb_+^{(1)} & \widehat\gb_-^{(1)} & \cdots &\widehat\gb_+^{(P)} & \widehat\gb_-^{(P)} \end{bmatrix} \in \Cd^{M\times 2P}$
	is also low-ranked, thanks to the the inter-coil annihilating filter relationship  \cite{jin2016general,Lee2016reference}.
	In \eqref{eq:Gb},  the superscript denotes the coil index.

	Therefore, if some of $k$-space data  are missing,
	the missing elements can be recovered using low rank Hankel matrix completion approaches \cite{jin2016general,ye2017compressive,lee2016acceleration,Lee2016reference,candes2009exact,cai2010singular,candes2010power,gross2011recovering,keshavan2010matrix}:
	\begin{eqnarray}\label{eq:EMaC}
	(MC)
	&\min\limits_{\widehat \Zb\in \Cd^{M \times 2P}} & \rank~ \hank_{d|2P} (\widehat \Zb)  \\
	&\mbox{subject to } & \Pc_{\Lambda_e}[\widehat\gb_+^{(i)}] = \Pc_{\Lambda_e}[\widehat \zb_{2i-1}]  , \\
	&& \Pc_{\Lambda_o}[\widehat\gb_-^{(i)}] = \Pc_{\Lambda_o}[\widehat \zb_{2i}]  ,\quad i=1,\cdots, P \nonumber 
	\end{eqnarray}
	where $\widehat\zb_i$ denotes the $i$-th column of $\widehat\Zb$, $\Lambda_e$ and $\Lambda_o$ denote the $k$-space indices for the even and odd phase encodings, and $\Pc_\Lambda$ denotes the projection to the sampled index $\Lambda$. After solving $(MC)$,  we can obtain the interpolated even and odd virtual $k$-space data for each coil, from which the virtual even and odd images are obtained for each coil by simply taking the inverse Fourier transform.
	Then, the final artifact corrected EPI data can be obtained as the sum of squares. This  procedure of ALOHA-based ghost removal is illustrated in Fig.~\ref{fig:overview}(a).

	The low-rank Hankel matrix completion problem $(MC)$ can be solved in various ways, and ALOHA employs
	the matrix factorization approaches  \cite{jin2016general,lee2016acceleration,Lee2016reference}. 
	The main technical problem, however, is the relatively expensive computation costs for the matrix factorization.
	Moreover, in order to improve the performance of ALOHA, the  matrix pencil size for Hankel matrix should  be sufficiently large, which in turn introduces significantly increased computational burden and memory size, and makes the algorithm unstable.
	In the following section, we show that a deep learning approach can address this problem by handling the matrix decomposition using a training data-based neural network.

	\subsection{From ALOHA to Deep Neural Network} 
	
	In order to explain the link between ALOHA and deep neural network, 
	here we first explain ALOHA as a data-specific basis representation, and  show that cascaded implementation of the
	ALOHA with ReLU nonlinearity makes it an encoder-decoder convolutional neural network  (CNN) architecture with
	good generalizability and expressivity.

	Specifically, suppose that a feasible solution  $\hat\Zb \in \Cd^{M\times 2P}$ for (MC)  has  the associated Hankel matrix
	$\hank_{d|2P}(\widehat\Zb) \in \Cd^{M\times 2dP}$ with a rank $Q$.
	Then, we can find the two set of basis matrices
	$\Psib, \widetilde{\Psib}\in \Rd^{2dP \times Q}$  and  $\Phib, \widetilde{\Phib} \in \Cd^{S \times M}$ with $Q\leq 2dP$ and $S\geq M$
	such that
	\begin{eqnarray}\label{eq:projection}
	\Psib \widetilde \Psib^{\top} = \Pb_{Row}&,& \Phib \widetilde \Phib^{\top} = \Ib_{M},
	\end{eqnarray}
	where $^\top$ denotes the Hermitian transpose,
	$\Pb_{Row}$ denotes the projection matrix to the row subspace of $\hank_{d|2P}(\hat \Zb)$ and $\Ib_M$ refers to the $M\times M$ identity matrix.
	Using these matrices, it is easy to see that 
	the Hankel matrix $\hank_{d|2P}(\widehat\Zb)$ has the following rank-$Q$ decomposition:
	\begin{eqnarray}
	\hank_{d|2P}\left(\widehat \Zb \right) &=& \widetilde\Phib \Phib^{\top}\hank_{d|2P}\left(\widehat \Zb \right) \Psib \tilde \Psib^{\top} =  \widetilde\Phib \Cb \tilde \Psib^{\top} \label{eq:Hdec} 
	\end{eqnarray}
	where  the expansion coefficient  $\Cb \in \Cd^{S\times Q}$, which is often called the convolution framelet coefficients \cite{ye2018deep},
	is given by
	\begin{eqnarray}\label{eq:Henc}
	\Cb := \Phib^{\top}\hank_{d|2P}\left(\widehat\Zb\right) \Psib   \quad \in \Cd^{S\times Q}
	\end{eqnarray}

	One of the most important findings in  the mathematical theory of {\em deep convolutional framelets} \cite{ye2018deep} is that
	Eq.~\eqref{eq:Hdec} can be represented  as a multi-channel convolution \cite{ye2018deep}:
	\begin{eqnarray}\label{eq:dec}
	\widehat\Zb= \left(\widetilde\Phib\Cb \right) \circledast g(\widetilde\Psib)
	\end{eqnarray}
	where $\circledast$ is the convolution, and $g(\widetilde\Psib)$ denotes multi-channel filters that are obtained by rearranging components of $\widetilde\Psib$.
	Similarly,  \eqref{eq:Henc} can be represented by another multi-channel convolution   \cite{ye2018deep}:
	\begin{eqnarray}\label{eq:enc}
	\Cb = \Phib^\top \left( \widehat \Zb \circledast h(\Psib) \right),
	\end{eqnarray}
	where $ h(\Psib)$ is the multichannel filter obtained by rearranging $\Psib$. 
	Interestingly, the order of the operation in \eqref{eq:enc}, which is composed of multi-channel convolution followed by global matrix multiplication, has striking similarity with the encoder neural network architecture that is composed of multi-channel convolution followed by pooling operation.
	Additionally,  we can find the similarity between \eqref{eq:dec} and the decoder neural network architecture that is composed of unpooling and multichannel convolution.
	Thus, $\Phib^\top$ and $\tilde\Phib$ in \eqref{eq:enc} and \eqref{eq:dec} are identified as the generalized pooling and unpooling operation \cite{ye2018deep}, respectively; and Eqs.~\eqref{eq:enc} and \eqref{eq:dec} indeed constitute a one layer encoder-decoder CNN (E-D CNN) architecture.  For more details, see \cite{ye2018deep}.

	However, the perfect reconstruction condition used in \eqref{eq:Hdec} is not interesting in neural networks, since in our $k$-space interpolation problem the input of the neural network is sub-sampled $k$-space data, whereas the output should be interpolated full $k$-space data. Furthermore, the representation should be generalized well for various $k$-space EPI input data rather than specific input at the training phase. To address this problem, the authors in \cite{ye2019understanding} showed that the cascaded implementation of \eqref{eq:enc} and \eqref{eq:dec} 
	with ReLU nonlinearity results in an E-D CNN that generalizes well and has exponentially large number of representation with respect to depth, channel, and skipped connections.
	
\begin{figure}[h]
		\centering
		\includegraphics[width=8cm]{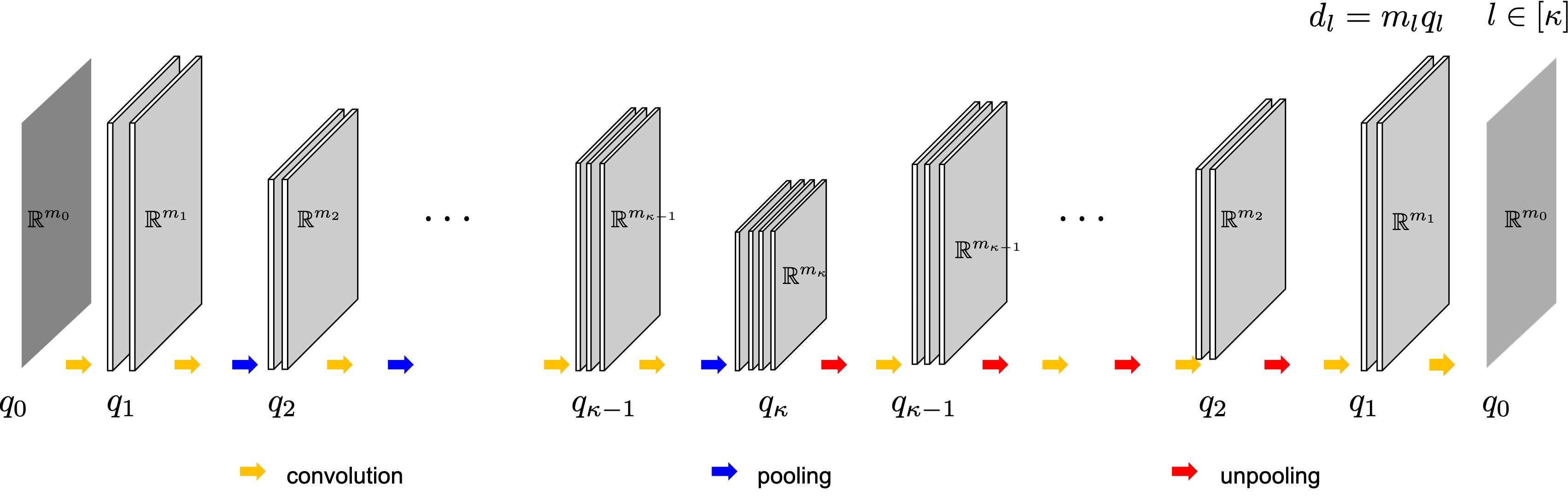}
		\caption{
			Simplified encoder-decoder CNN architecture. At the $l$-th layer, $m_l$, $q_l$ and $d_l:=m_lq_l$ denote the dimension of the signal,  the number of filter channel, and the	total feature vector dimension, respectively. We consider symmetric configuration  so that both encoder and decoder have the same number of layers, say $\kappa$.
		}
		\label{fig:geometry}
	\end{figure}

	To understand this claim, consider a cascaded version of the E-D CNN with rectified linear unit (ReLU) nonlinearities
	as shown in Fig.~\ref{fig:geometry}. By vectorizing \eqref{eq:enc} and \eqref{eq:dec} with the superscript $l$ referring to the $l$-th layer, the authors in \cite{ye2019understanding} showed that the $l$-th layer encoder and decoder  can be represented  by:
	\begin{eqnarray}\label{eq:ED}
	\z^l=\sigma(\Eb^{l\top} \z^{l-1})&,& \tilde \z^{l-1}=\sigma(\Db^l \tilde\z^{l})
	\end{eqnarray}
	where $\sigma(\cdot)$ denotes the element-by-element ReLU, and 
	$\z^l$ is a vectorized version of the  $l$-th layer signal $\Z^l$,  
	and the encoder and decoder matrices are defined by
	\begin{eqnarray}\label{eq:El}
	\E^l= \begin{bmatrix} 
	\Phib^l\circledast \psib^l_{1,1} & \cdots &  \Phib^l\circledast \psib^l_{q_l,1}  \\
	\vdots & \ddots & \vdots \\
	\Phib^l\circledast \psib^l_{1,q_{l-1}} & \cdots &  \Phib^l\circledast \psib^l_{q_{l},q_{l-1}}
	\end{bmatrix}
	&,&
	\Db^l= \begin{bmatrix} 
	\tilde\Phib^l\circledast \tilde\psib^l_{1,1} & \cdots &  \tilde\Phib^l\circledast \tilde\psib^l_{1,q_l}  \\
	\vdots & \ddots & \vdots \\
	\tilde\Phib^l\circledast \tilde\psib^l_{q_{l-1},1} & \cdots &  \tilde\Phib^l\circledast \tilde\psib^l_{q_{l-1},q_{l}}
	\end{bmatrix}
	\end{eqnarray}
	and
	\begin{eqnarray*}
		\begin{bmatrix} \Phib^l \circledast \psib_{i,j}^l  \end{bmatrix}  :=\begin{bmatrix} \phib^l_1 \circledast \psib_{i,j}^l & \cdots & \phib^l_{m_l} \circledast  \psib_{i,j}^l\end{bmatrix}  \label{eq:defconv}
	\end{eqnarray*}
	%
	%
	Then, the output $\yb$ of the multi-layer encoder-decoder CNN (ED-CNN)  with respect to input $\z$ can be represented by nonlinear frame representation  \cite{ye2019understanding}:
	\begin{eqnarray}\label{eq:basis}
	\yb 
	~=  \sum_{i} \left\langle {\blmath b}_i(\z), \z \right\rangle \tilde  {\blmath b}_i(\z) ~:= \Kc\left(\z; \left\{\Psib^l,\tilde\Psib^l\right\}_{l=1}^\kappa\right)
	\end{eqnarray}
	where $ {\blmath b}_i(\z)$ and $\tilde  {\blmath b}_i(\z)$ denote the $i$-th column of the following frame basis and its dual, respectively:
	\begin{eqnarray}
	\B(\z)&=& \Eb^1\Lambdab^1(\z)\Eb^2 \cdots  \Lambdab^{\kappa-1}(\z)\Eb^{\kappa},~\quad \label{eq:Bc}\\
	\tilde \B(\z) &=& \Db^1\tilde\Lambdab^1(\z)\Db^2 \cdots  \tilde\Lambdab^{\kappa-1}(\z)\Db^{\kappa} \label{eq:tBc}
	\end{eqnarray}
	where
	$\Lambdab^l(\z)$ and $\tilde\Lambdab^l(\z)$ denote the diagonal matrix with 0 and 1 values that are determined by the ReLU output
	in the previous convolution steps \cite{ye2019understanding}.
	
	Now, the nonlinear frame basis ${\blmath b}_i(\z)$ and its dual $\tilde  {\blmath b}_i(\z)$ in \eqref{eq:basis} have explicit dependency   on the input $\z$  in \eqref{eq:basis}, due to the ReLU output
	$\Lambdab^l(\z)$ and $\tilde\Lambdab^l(\z)$ in    \eqref{eq:Bc} and \eqref{eq:tBc}. This makes the signal representation via basis adaptively vary  depending on input signals.  In fact, thanks to the combinatorial nature of ReLU in  \eqref{eq:Bc} and \eqref{eq:tBc}, the number of distinct
	linear representation increases exponentially with the depth, width and the skipped
	connection of the network, in spite of using the same set of learned filters \cite{ye2019understanding}.
	This is believed to be the  main origin of the superior performance of deep neural networks.

	\subsection{Interpolation Kernel Estimation using CNN}
	
	Based on the above discussion, the network training problem  can be formulated as follows:
	\begin{eqnarray}\label{eq:newcost}
	\min_{ \{\Psib^l, \widetilde\Psib^l\}_{l=1}^\kappa} \sum_{i=1}^P
	\left\|  \begin{bmatrix} \widehat \gb_+^{(i)} \\ \widehat \gb_-^{(i)} \end{bmatrix} -
	\Kc\left( \begin{bmatrix} \Pc_{\Lambda_e}\left[\widehat \gb_+^{(i)} \right]\\  \Pc_{\Lambda_o}\left[\widehat \gb_-^{(i)} \right]\end{bmatrix}, \left\{\Psib^l,\tilde\Psib^l\right\}_{l=1}^\kappa\right)
	\right\|^2 
	\end{eqnarray}
	where $\widehat \gb_+^{(i)}$ and $\widehat \gb_-^{(i)}$ are fully sample virtual even and odd $k$-space data,
	and $\Pc_{\Lambda_e}$ and $\Pc_{\Lambda_o}$ are projection to the even and odd phase encoding lines, respectively.
	Thus, the neural network problem is essentially to find the $k$-space interpolation kernel to estimate the missing $k$-space data. Therefore, if the EPI data is also accelerated, the input data index $\Lambda_e$ and $\Lambda_o$ have more zeros due to the sub-sampling; other than this, the neural network training problem is essentially the same.

	In the $k$-space deep learning for accelerated MRI \cite{han2018k}, it is recommended that the network is implemented in the $k$-space domain, while the image domain loss is minimized. If $l_2$-loss is used, this is equivalent to \eqref{eq:newcost}, but when other sophisticated image domain properties are exploited, it is usually recommended to re-define a loss in the image domain. In spite of this hybrid architecture, the gradient can be easily calculated, since the adjoint of the Fourier transform is the inverse Fourier transform that can be easily implemented using the fast Fourier transform (FFT) \cite{han2018k}.
	Therefore,  by following the same idea of the $k$-space deep learning \cite{han2018k}, in the proposed algorithm, ghost corrected images are used as for the label data for network training with the $l_2$ loss, while the network is implemented in the $k$-space.
	
	However, care should be taken in the implementation since the input $k$-space data are {\em virtual}  even and odd  $k$-space data. Thus, we should provide reconstruction images from the interpolated {\em virtual} $k$-space data as labels. Since ALOHA-based ghost correction can interpolate the virtual even and odd $k$-data, we therefore use the reconstruction results from these interpolated $k$-space data as our label.  Specifically, as shown in Fig.~\ref{fig:overview}(b), the even and odd virtual images are generated using ALOHA for each coil from ghost corrected even and odd $k$-space lines.   Then, these even and odd complex images for each coils  are used as label data.

	\section{Method}

	In this work, we used 3T and 7T GRE (gradient-recalled-echo) EPI data. The 3T raw data were acquired by Phillips 3T whole body MR scanner (Phillips Achieva system) for 17 healthy volunteers using 8 channel phased-array RF coils. Acquisition parameters were as follows: TR/TE = 3000/30 $ms$, 36 slices with 3 $mm$ slice thickness, Field of View (FOV) = 240$\times$240 $mm^2$, voxel size = $3.75\times3.75 mm^2$, and 60 temporal frames acquired with full Fourier sampling. The 7T raw data were acquired by 7T whole body MR scanner (Siemens Magnetom Terra) for 7 healthy volunteers using 32 channel phased-array RF coils. The acquisition parameters were as follows: TR/TE = $5800/55 ms$, Field of View (FOV) = $196\times196 mm^2$, voxel size = $1\times1 mm^2$, slice thickness = $1 mm$. The acquired data is composed of 50 slices and 5 temporal frames. To acquire high spatial resolution, the 7T data were accelerated by a factor of 2.
	
	The pre-scan data for conventional ghost correction method is also acquired before the whole data acquisition. For both 3T and 7T, this pre-scan data is acquired for all slices and coils without phase encoding blips. The phase difference is calculated by subtracting a phase of one PE line from a phase of adjacent PE line. Then the phase mismatch of image is compensated by the calculated phase difference map. 
	Also, the reference-free ghost correction algorithm in \cite{skare2006afast} is implemented. 
	In this method, the phase difference is assumed to be a linear function, and a correction value that minimizes the entropy is found. We also implemented PEC-SENSE \cite{xie2018robust} and ALOHA for comparative studies.  For the PEC-SENSE implementation, we obtained the coil sensitivity map by using the ESPIRiT algorithm \cite{Uecker2014espirit} from the ghost-corrected multi-channel images  that are obtained using a ALOHA based correction method. This
	is because PEC-SENSE requires accurate   coil sensitivity map, and as ALOHA provides the most accurate reconstruction results for
	coil sensitivity map estimation. All these methods were implemented on the same computing hardware as the proposed deep learning approach (GTX 1080-Ti graphic processor and i7-4770 CPU (3.40GHz)).
	
\begin{figure}[h]
		\centering
		\includegraphics[width=12cm]{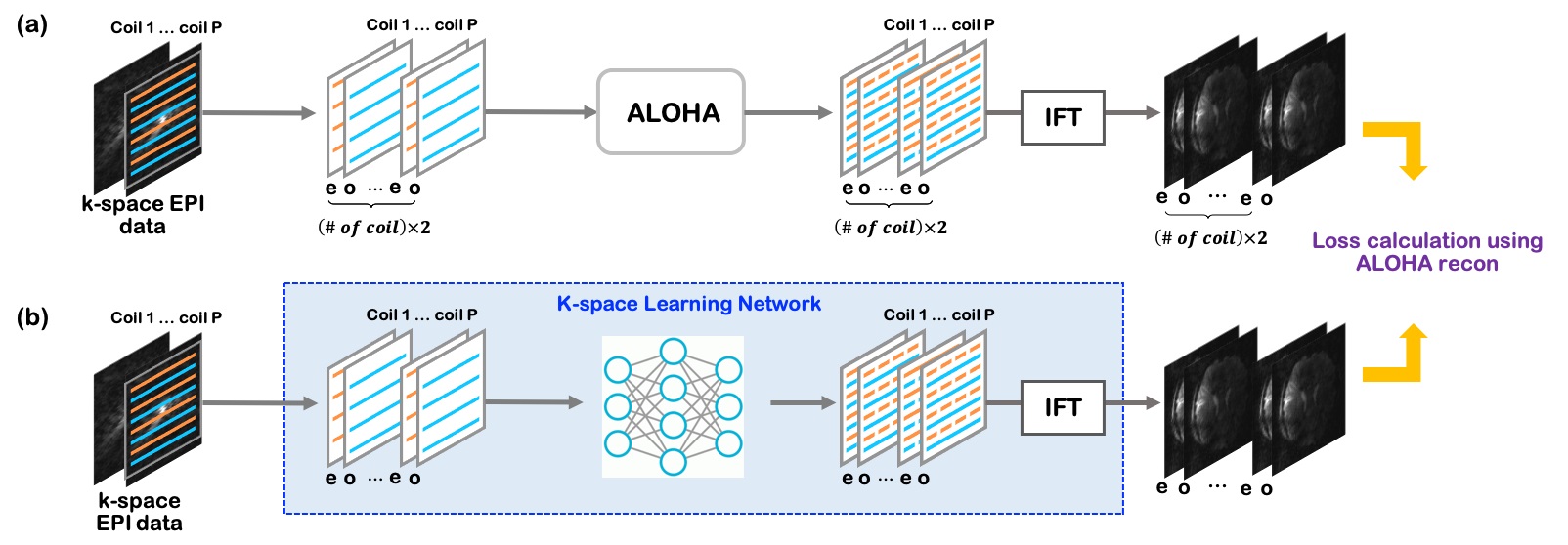}
		\caption{
			Reconstruction flow for (a) ALOHA-based ghost artifact removal, and (b) the proposed $k$-space deep learning for ghost artifact removal. Here, $e$ and $o$ refer to the frames composed of the even and odd $k$-space lines, respectively. After the interpolated even and odd virtual images are generated, the sum-of-squares image is obtained as the final ghost corrected image. Here, IFT stands for the inverse Fourier transform. }
		\label{fig:overview}
	\end{figure}

	The overall reconstruction flow of the proposed $k$-space deep learning-based ghost correction method  is illustrated in Fig.~\ref{fig:overview}(b). 
	Specifically, the $ P $ channel multi-coil $ k $ space data is divided into even and odd channels. Then, the even and odd $k$-space data are processed using a E-D CNN, whose output is the interpolated even and odd $k$-space data for each coil.
	Then, the inverse Fourier transform is applied to obtain the ghost corrected virtual even and odd images.
	Finally, similar to ALOHA,  the sum-of-squares images are generated from the even and odd images from multiple channels. In fact, this procedure is exactly the same at the ALOHA-based ghost correction algorithm in Fig.~\ref{fig:overview}(a), except for the $k$-space interpolation step. Specifically, $k$-space interpolation is done using ALOHA in Fig.~\ref{fig:overview}(a), whereas it is done using a E-D CNN in  Fig.~\ref{fig:overview}(b).
	Another difference is that ALOHA uses the zero-padded even and odd $k$-space data as input, whereas in the proposed method we initialized the missing $k$-space lines with the other phase $k$-space data.
	More specifically, for the virtual even $k$-space missing lines (resp. for the virtual odd $k$-space missing lines),
	we initialize the missing lines with the measured odd $k$-space lines (resp. measured  even $k$-space lines).  Because all these initialization data comes from the measurement from the identical slice at the identical temporal frame, there are no additional information added. We found that this initialization make the network training converge faster than just zeroing-out the signals.

	Fig.~\ref{fig:twomethod}(a) shows the proposed direct training scheme for the accelerated EPI acquisition
	for the case of our 7T data.  In this case, the label data is phase-corrected full $k$-space data, so that the neural
	network is trained to learn both phase correction and interpolation kernel. As an alternative method,  Fig.~\ref{fig:twomethod}(b) shows  a hybrid scheme, where half-ROI images from phase-corrected  and downsampled EPI $k$-space data  were used as our label for the neural network training, after which final $k$-space data is generated by GRAPPA kernel estimation using calibration data. Unlike the method in Fig.~\ref{fig:twomethod}(a), the method in Fig.~\ref{fig:twomethod}(b) requires additional computation of GRAPPA kernel using the
	calibration data for each case. The main motivation for implementing the alternative method in Fig.~\ref{fig:twomethod}(b) is to show that the proposed training method can estimate the interpolation kernels for both GRAPPA and phase errors so that additional calibration data is not necessary.
	
	Recall that we used the ALOHA-based ghost correction images as label. However, as will be shown later, for high-field MRI, the ALOHA-based correction algorithm often fail to remove the ghost artifacts in some frames. Therefore,  we visually inspected ALOHA-based ghost correction results, chose the ones with ghost corrected $k$-space data, and use them as labels. Interestingly, as will be shown later, our trained neural network then successfully corrects the ghost artifacts even from the cases where ALOHA fails. 
	Also, to make a robust neural network which can correct non-linear phase error, we additionally used  simulated input data with non-linear phase error. The ratio of the original ghost input to the non-linear phase error simulation input was 50:50.

\begin{figure}[h]
		\centering
		\includegraphics[width=10cm]{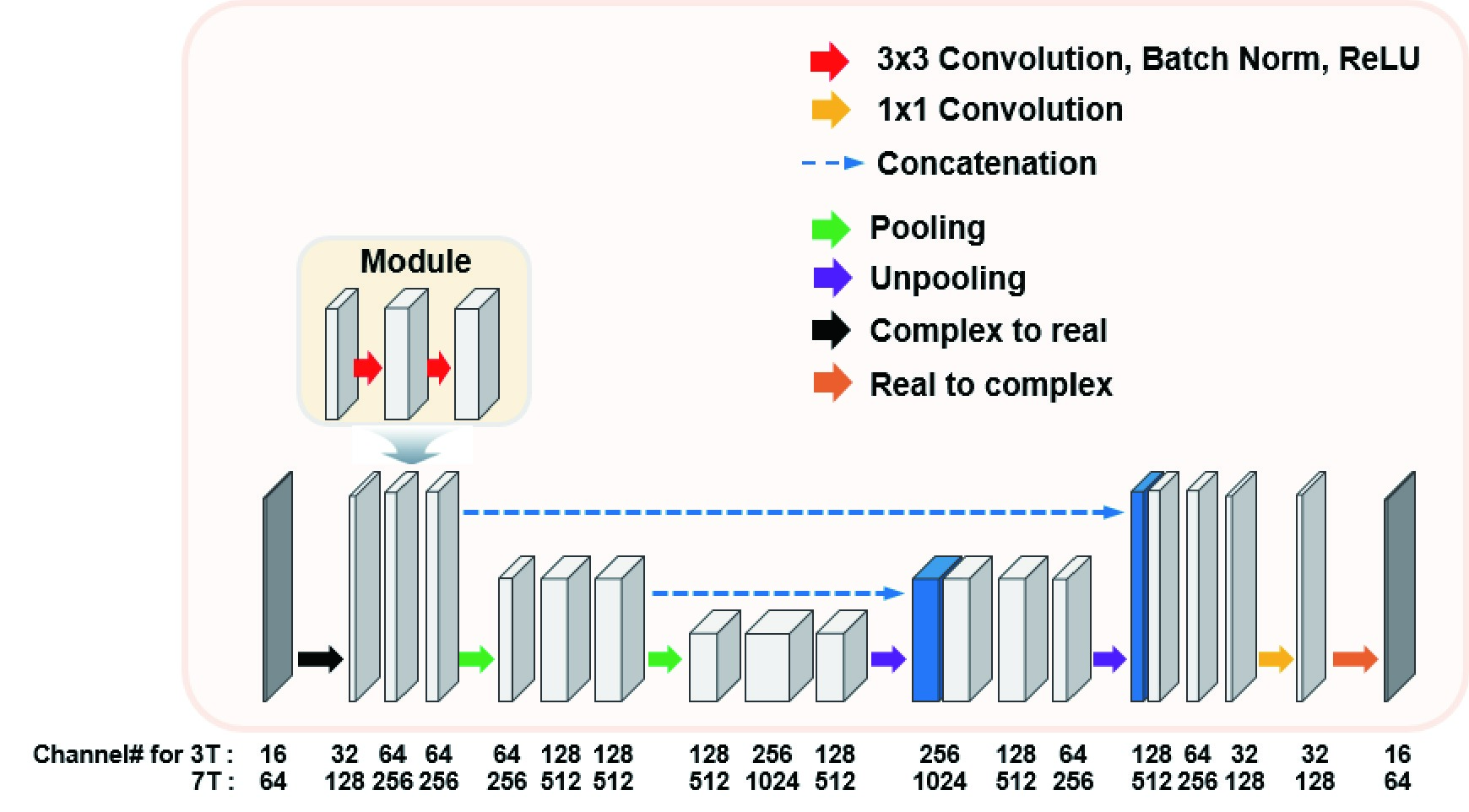}
		\caption{
			The proposed k-space learning network backbone for the EPI data ghost correction. This neural network is composed of convolution layers with batch normalization and ReLU, pooling layer, and contracting path with concatenation. Here, red arrow refers to several 3$\times$3 convolution layers with batch normalization and rectified linear unit (ReLU). Also, 2$\times$2 average pooling (green arrow) and unpooling layers (purple arrow) are inserted after two basic unit layers. Some of intermediate image on encoding process is concatenated with unpooled image (blue arrow). The yellow arrow which is located at the end of network, indicates 1$\times$1 convolution. As the input k-space and output image data are complex-valued, so we first divide the complex-valued $k$-space data into real and imaginary channels(black arrow). Accordingly, the number of channels are doubled because this step separates real and imaginary value of $k$-space and concatenates them through channel direction. Similarly, at the network output, the real-valued real and imaginary channels are combined to obtain complex-valued $k$-space data (brown arrow).}
		\label{fig:structure}
	\end{figure}

	Fig.~\ref{fig:structure} illustrates a multi-scale neural network backbone employed in the proposed method. Specifically, the network backbone is based on U-net structure \cite{ronneberger2015u}  composed of convolution layers with batch normalization and ReLU, pooling layer, and contracting path with concatenation. Here, red arrow refers to the basic unit of this network that consists of several 3$\times$3 convolution layers with batch normalization and rectified linear unit (ReLU) after every convolution layer. Also, 2$\times$2 average pooling (green arrow) and unpooling layers (purple arrow) are inserted after two basic unit layers. Some of intermediate image on encoding process is concatenated with unpooled image (blue arrow). The yellow arrow which is located at the end of network, indicates 1$\times$1 convolution. 
	As the input k-space and output image data are complex-valued, so  we  first divide the complex-valued $k$-space data into real and imaginary channels as suggested in \cite{han2018k} before using them as inputs to U-Net backbone. This operator is shown in Fig.~\ref{fig:structure} as black arrow. Accordingly,
	the number of channels are doubled because this  step separates real and imaginary value of $k$-space and concatenates them through channel direction. Similarly, at the network output, the real-valued real and imaginary channels are combined to obtain complex-valued $k$-space data (brown arrow).

	The network was implemented using MatConvNet toolbox (ver.24) in MATLAB 2015a environment (Mathwork, Natick). We used a GTX 1080-Ti graphic processor and i7-4770 CPU (3.40GHz). The weights of convolutional layers were initialized by Gaussian random distribution with Xavier method to achieve proper scale. This helps us to avoid the signal to be exploded or vanished in the initial phase of learning. The stochastic gradient descent (SGD) method with momentum was used to train the weights of the network and minimized the loss function. Training was conducted until the learning curve converged. The learning rate was started from $10^{-4}$, and gradually decreased to $10^{-6}$. To prevent the weight becoming too large, the regularization parameter is used as $10^{-4}$ for both 3T and 7T EPI training. Also, we used $l_2$ loss of image domain data for training. Training time was 4 hours and 63 hours for 3T and 7T, respectively. For the case of 3T data,
	the number of subjects are 17, among which 14 subjects are used for training, 2 subjects are used as validation, and the other subject is used as test. In 7T training, the number of subjects are 7, so 5 subjects are used for training, 1 subject is used as validation, and the other is used as test.

	To verify the advantages of proposed algorithm for functional MRI, the SPM (Well-come Trust Centre for Neuroimaging, London, UK) analysis is performed using corrected results from 60 temporal frames from 3T MR system. The EPI data were acquired with a pair hand squeezing stimulation. Total acquisition time was 180s, and the hand squeezing tasks were performed between 30-60s, 90-120s, and 150-180s. 
	Standard spatial preprocessing was performed before the general linear model analysis. First, the images were spatially realigned to remove motion artifacts in time series data. Then, a co-registration process between the motion corrected EPI image and the T1 source image was conducted using a voxel to voxel affine transformation matrix. Smoothing with a $6mm \times 6mm \times 6mm$ full-width at half maximum Gaussian kernel was applied to the images. Statistical analysis was conducted based on the general linear model for the spatially preprocessed images. To model the hemodynamic response, the canonical hemodynamic response function was used. After the specification of the response model and the design matrix, parameter estimation was performed using restricted maximum likelihood. For statistical analysis, a family-wise error corrected significance threshold of 5\% was used ($P < 0.05$).

	\section{Results}
	
\begin{figure}[h]
		\centering
		\includegraphics[width=11cm]{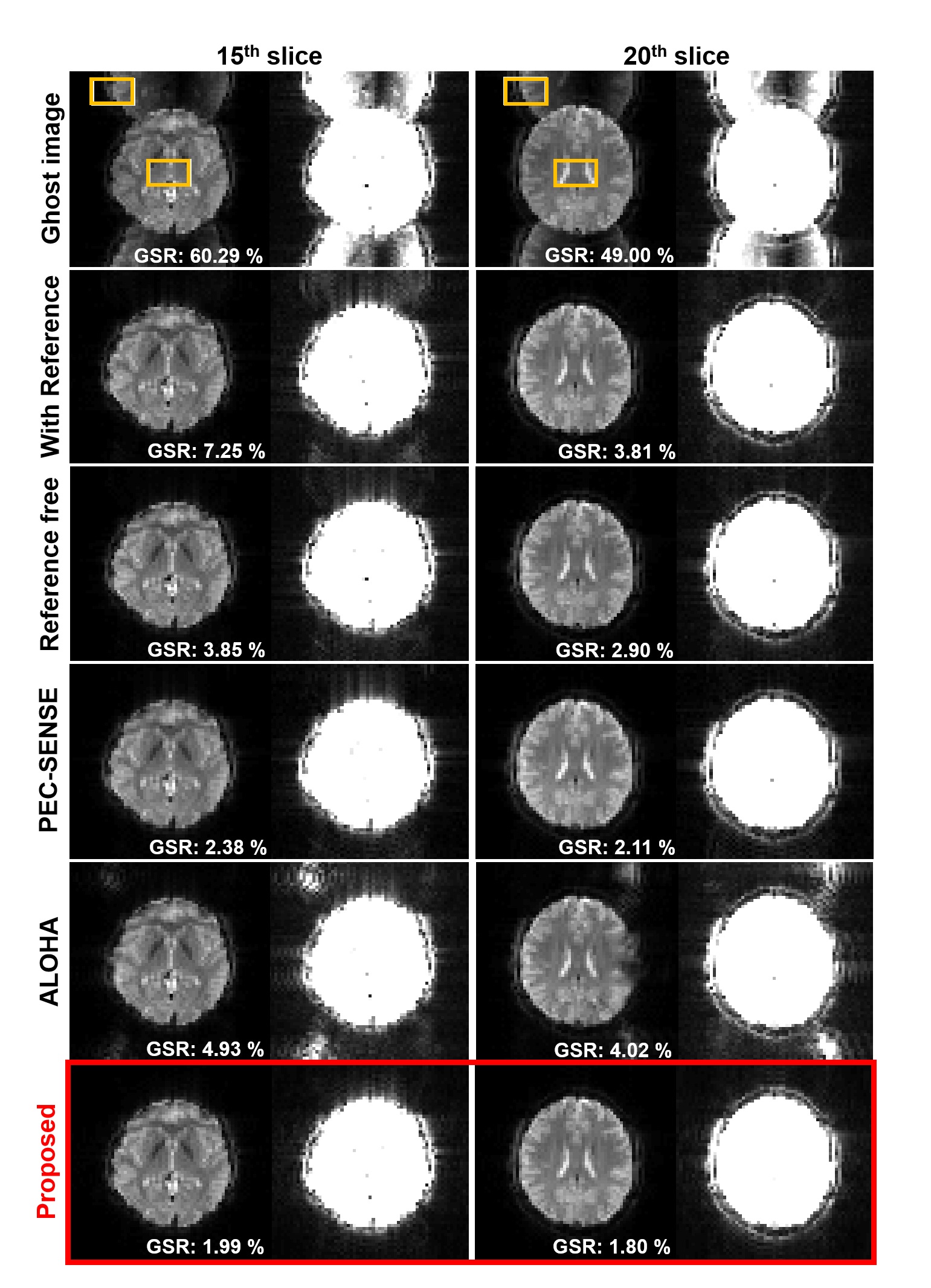}
		\caption{
			Ghost correction results of 3T GRE-EPI in vivo data by various methods. Multi coil data is used to remove artifact. For each slice, the 2nd column shows the intensity rescaled image of 5 times. In the rescaled images, there are still remaining ghost and streaking artifacts in the other methods except the proposed method. ROI for calculating signal and ghost values are depicted by yellow rectangular.}
		\label{fig:result_3T}
	\end{figure}

	The results for 3T EPI ghost correction is shown in Fig.~\ref{fig:result_3T}. The intensity re-scaled result images are also shown in the figure. In order to observe the remaining ghost more concisely, the image scale was re-scaled at 5 times of original intensity. The  reconstruction performance of various ghost correction methods are mostly good in 3T EPI data, but  slightly remaining ghost artifact can be seen in the intensity magnified images from the conventional methods.  
	Now, Fig.~\ref{fig:result_7T} shows the 7T EPI ghost correction result with corresponding 5 times increase in brightness image. Most correction methods except the proposed method shows a remaining ghost artifact. 
	Furthermore, Fig.~\ref{fig:result_7T} confirmed that
	the proposed method for accelerated EPI data (see Fig.~\ref{fig:twomethod}(a))
	can accurately correct ghost artifacts and provide visually similar results from the network in Fig.~\ref{fig:twomethod}(b) that explicitly utilizes the
	GRAPPA kernel estimation steps.

	To compare the result quantitatively, we calculate the ghost to signal ratio and mark ghost-to-signal ratio (GSR) value on the result figure. The GSR values corresponding to each result are calculated using regions represented by the yellow boxes and marked on the image. As shown in this result, the proposed method successfully removed ghost artifacts and shows the lowest GSR value. Although the performance of the PEC-SENSE which uses the parallel imaging information is better than the other existing methods, the GSR value was higher than the proposed method.  Moreover,
	the proposed method using Fig.~\ref{fig:twomethod}(a) provides much smaller GSR values than the network training using
	Fig.~\ref{fig:twomethod}(b).

\begin{figure}[h]
		\centering
		\includegraphics[width=12cm]{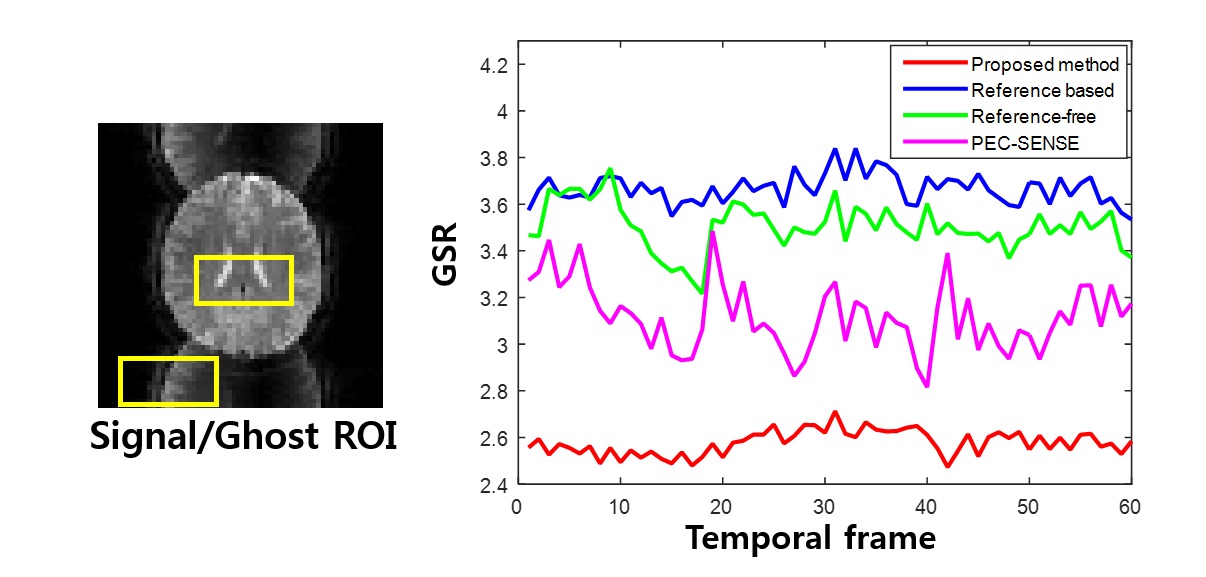}
		\caption{
			Temporal GSR plots for various ghost artifact correction methods for the case of 3T EPI data. To get the GSR, signal value is calculated by the yellow box inside brain region, and ghost value is calculated outside the brain. The proposed method shows the temporal stability in its performance to correct ghost artifacts.}
		\label{fig:temporalGSR}
	\end{figure}

	Fig.~\ref{fig:temporalGSR} shows the temporal GSR values of the proposed method and other correction methods for the case of 3T. 
	The GSR values were calculated on 60 temporal frames. The ROI for calculating signal value is marked on the brain image as yellow box. Ghost value is calculated on the outside of brain region. As shown in Fig.~\ref{fig:temporalGSR}, among the existing methods,
	PEC-SENSE was best, but the proposed method significantly outperform PEC-SENSE and provided
	consistently best GSR compared to the other methods. 
	
\begin{figure}[h]
		\centering
		\includegraphics[width=14cm]{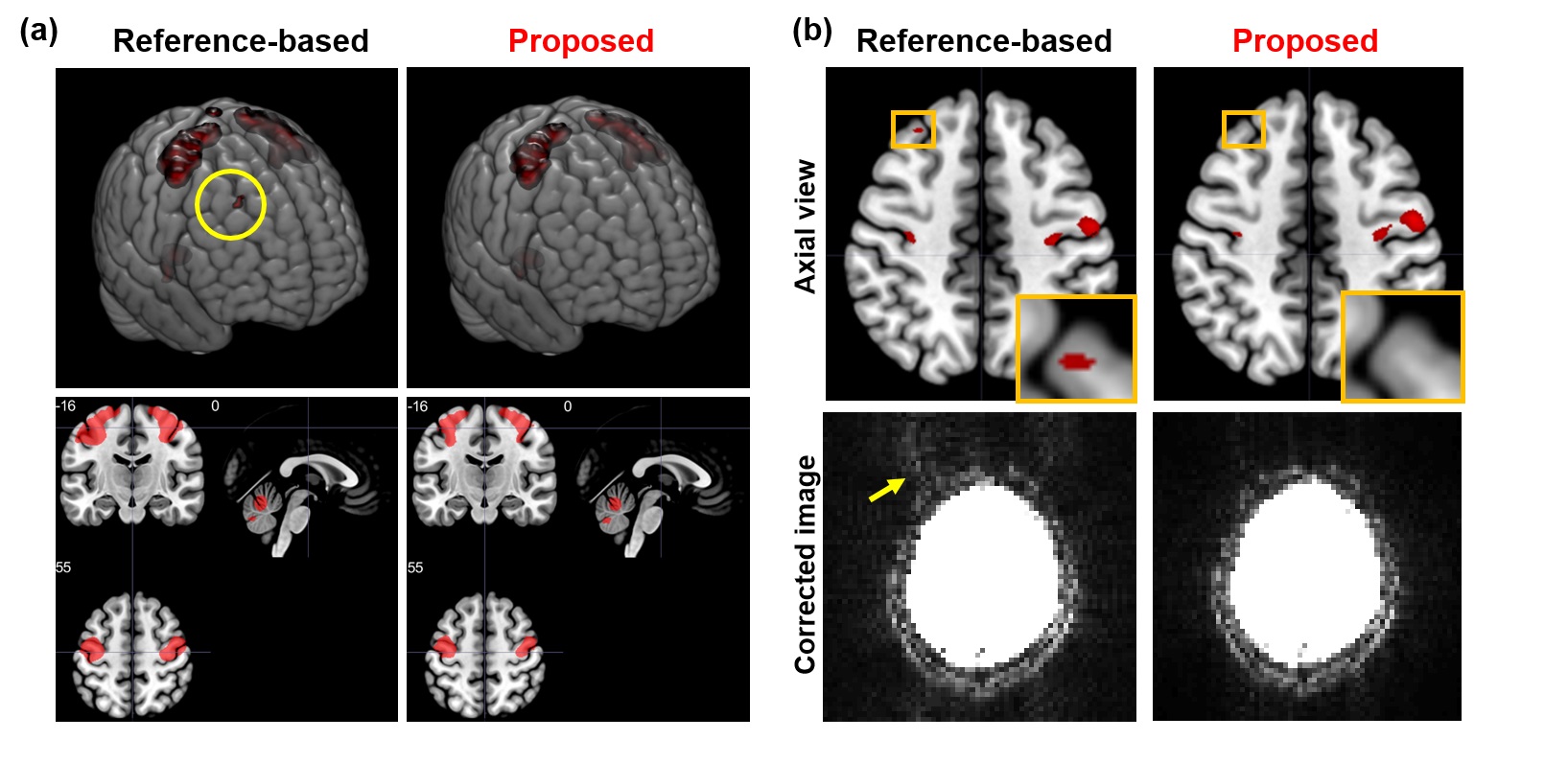}
		\caption{
			(a) SPM analysis result of 3T GRE-EPI data, which is visualized by MRIcroGL software. The proposed method shows the activation on motor cortex regions only, while the reference-based method shows suspicious activation. (b) The reconstructed slice at the artifact locations showed that there are still remaining	ghost artifacts for the conventional methods, which is not the case for the	proposed method. }
		\label{fig:spm_result}
	\end{figure}

	Additionally,  SPM analysis was conducted to demonstrate the potential errors in the
	functional activity estimation from the artifact corrected data. The activation of corrected multi-coil 3T EPI data is calculated by SPM, and visualized by MRIcroGL (http://www.cabiatl.com/mricrogl/). Pair hand squeezing stimulation shows a motor cortex activation on brain. Figs.~\ref{fig:spm_result}(a)  show the SPM results of reference-based and proposed method. There is a suspicious activation on the prefrontal area when reference-based method was used. Figs.~\ref{fig:spm_result}(b) show  the magnified axial view and corresponding reconstruction slice of the suspicious activation. As marked by yellow arrow on the reconstructed image, there is a remaining ghost artifact on the reference-based method. Since this artifact is only shown on the reference-based method result, we conjecture that this artifact is likely to have affected the signal in the pre-frontal area of the brain, which may have influenced functional analysis as well.

	We compared the computational time for various methods. The computational time for ghost correction was calculated for each single slice and single temporal frame with multi coil $k$-space data.
	It is also important to note that at the inference stage, our neural network approaches do not need any ALOHA data since the trained neural network can directly estimate the missing k-space data as shown in Fig.~\ref{fig:overview}(c)(d). 
	Therefore, the run time complexity of our method is very small.
	For example, for 3T EPI ghost correction, the reference based method and the proposed method took 2.1$ms$,and 16.9$ms$, respectively. On the other hand, ALOHA took about 48.6sec, which is several order of magnitude slower than the proposed method. This confirms the practicability of the proposed method.

	\section{Discussion}  
	
	The proposed method outperforms the existing methods in both 3T and 7T EPI data.  While other conventional correction algorithms require additional scan or calibration data, our algorithm shows good performance to correct artifact without any additional scan. 
	One may wonder that training with multiple subject is necessary for each scan. However, this is not the case. For example, in our experiments with 7T data, once the neural network is trained with a few subjects, the same neural network can be consistently used for other subjects all the time unless the acquisition parameters are significantly changed.  This generalizability comes from the nature of the deep neural network.

\begin{figure}[h]
		\centering
		\includegraphics[width=10cm]{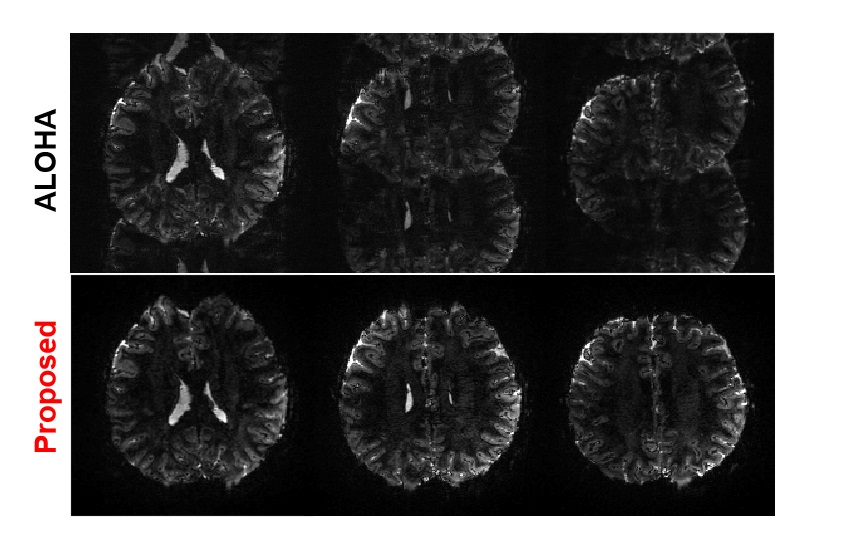}
		\caption{Robustness of the proposed method. Even for the failure cases by ALOHA, the proposed method successfully corrected the ghost artifacts, and the proposed method was successful for all experimental cases.}
		\label{fig:failed_edit}
	\end{figure}

	Recall that 7T EPI data is quite sensitive to the local field inhomogeneity, and the phase change is also large and nonlinear \cite{Hoge2016dual}. Therefore, the reference-based method which obtains a pre-scan from MR is often not able to perform the phase correction. Unlike the 3T EPI data, we found that ALOHA is not robust for 7T EPI due to the larger local field inhomogeneity variations. One could increase the matrix pencil size for the Hankel matrix to improve the performance of ALOHA, but the associated memory and computational complexity increases are sometime not acceptable in real applications. Moreover, with bigger matrix pencil size, the algorithm became more sensitive to other hyperparameters. Thus, ghost correction using ALOHA on 7T MR shows many failure cases. Specifically, from 6,250 set of slices in our experiment, we could get the only 1,346 successful cases for  ghost correction using the ALOHA. Thus, only 1,346 successful cases are used as label data for our neural network training.
	Fig.~\ref{fig:failed_edit} shows representative failure cases from the test data by the ALOHA-based ghost correction for 7T data. Interestingly, as shown in Fig.~\ref{fig:failed_edit}, even for the failure cases by ALOHA, the proposed method successfully corrected the ghost artifacts. In fact, among all data we used, no failure cases were observed using the proposed $k$-space deep learning approach for ghost correction. This shows that the proposed algorithm is very robust even under large local field inhomogeneity changes. Even in the cases where ALOHA was successful, the proposed method mostly outperforms ALOHA as shown in fig.~\ref{fig:result_3T} and fig.~\ref{fig:result_7T}. 
	Given that CNN is closely related to ALOHA, this phenomenon may appear mysterious.
	However, in CNN approaches, the resulting receptive field size from the U-Net covers all image, resulting in significantly large	matrix pencil size.  
	Moreover, as discussed before in detail, despite the use of the same filter coefficients, the number of distinct basis representation increases exponentially with the depth and width thank to the combinatorial nature of ReLU \cite{ye2019understanding}. This is in contrast to the data specific representation learning from Hankel matrix decomposition, which results in single linear representation. Therefore, the CNN approaches have significant advantages in terms of generalization capability and expressivity, which is not observed in the low-rank Hankel matrix decomposition.

\begin{figure}[h]
		\centering
		\includegraphics[width=12cm]{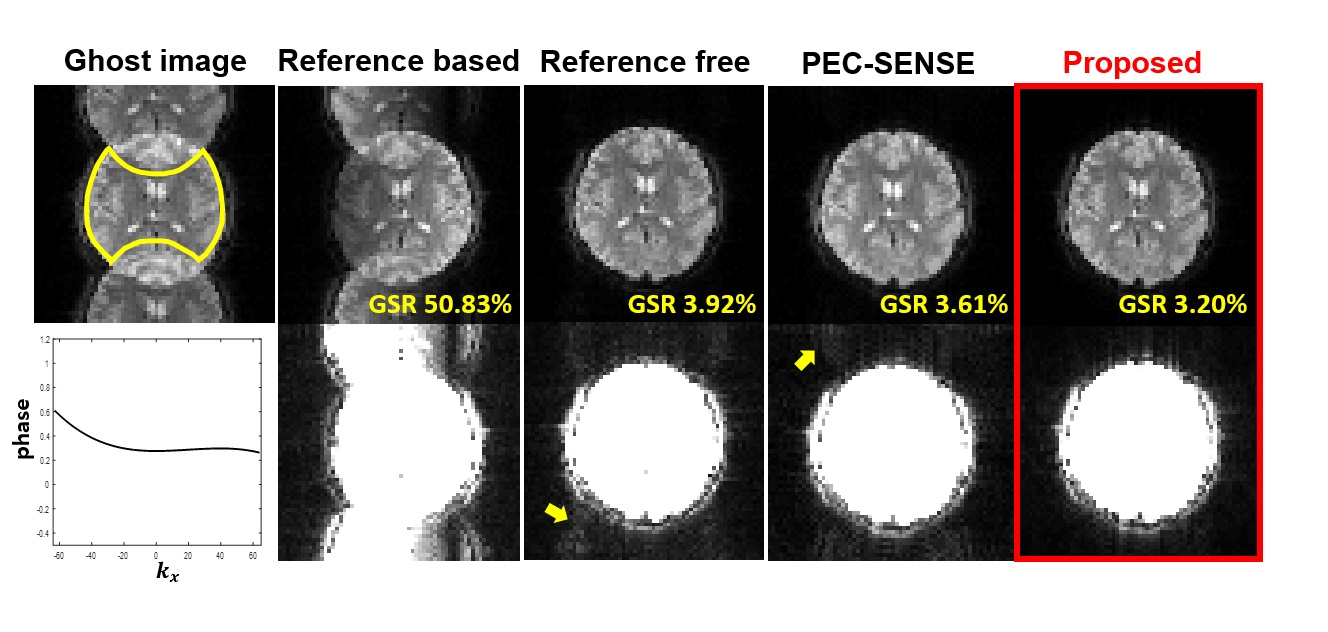}
		\caption{Ghost correction result for simulated non-linear phase error data. The data are based on 3T data and 
			the phase errors are modeled as the combination of cubic, quadratic, and constant error terms. Yellow boxes show the intensity rescaled images of 5 times. To get the GSR values, signal value is calculated inside the yellow line, and ghost value is calculated outside the brain. Here, the proposed method shows the most robust performance.
		}
		\label{fig:nonlinearphase}
	\end{figure}

	There are some ghost artifact correction methods which treats the non-linear phase disparity, such as DPG \cite{Hoge2016dual} and PEC-SENSE \cite{xie2018robust}. In order to correct the phase error of EPI precisely, DPG needs ghost-free EPI calibration data, and PEC-SENSE needs the  coil sensitivity maps. The proposed method does not need any additional acquisition, calibration data, and coil sensitivity maps. To show that the proposed method also corrects non-linear phase error, we  investigated the effects of simulated non-linear phase error data  using our proposed method and PEC-SENSE (similar to the studies in  \cite{Hoge2016dual}). Fig.~\ref{fig:nonlinearphase} shows non-linear phase error simulation (the combination of cubic, quadratic and constant error terms) and corresponding ghost correction result. The graphs show the phase error, and they are applied to ky-x projection domain. In the second row images that have magnified brightness 5 times, remaining ghost artifact are shown in the result of reference-free method and PEC-SENSE. On the other hand, the proposed method can correct non-linear phase errors without additional scan or calibration data. To compare quantitatively, we calculated ghost to signal ratio, which confirmed that our method was the best in correcting this nonlinear phase errors. This confirms the robustness of the proposed method for both nonlinear and linear phase perturbation data. More results for various non-linear phase error simulation are shown in Supporting information Figure S1. 
	
\begin{figure}[h]
		\centering
		\includegraphics[width=12cm]{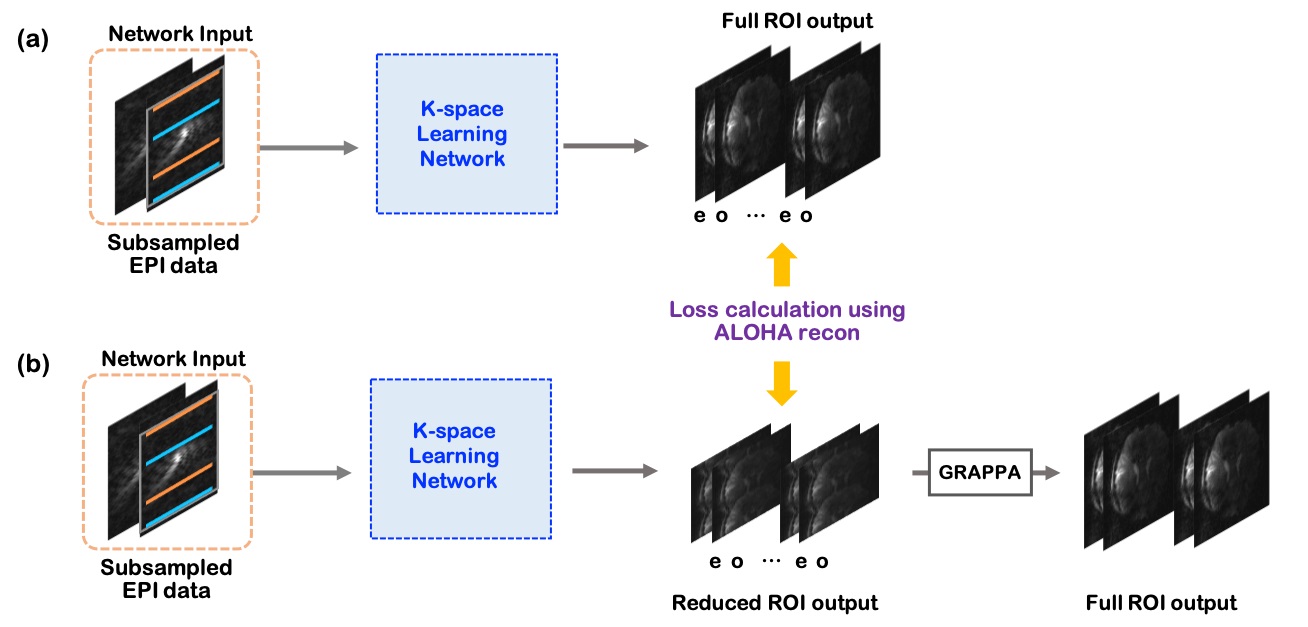}
		\caption{
			(a) Proposed neural network training process by using down-sampled EPI data as input and full ROI ALOHA images as label data. (b) A comparison network training process by using down-sampled EPI data as input and reduced ROI ALOHA images as label data. In this case, the final ghost corrected full ROI images are obtained by performing GRAPPA using calibration data.}
		\label{fig:twomethod}
	\end{figure}

\begin{figure}[h]
		\centering
		\includegraphics[width=13cm]{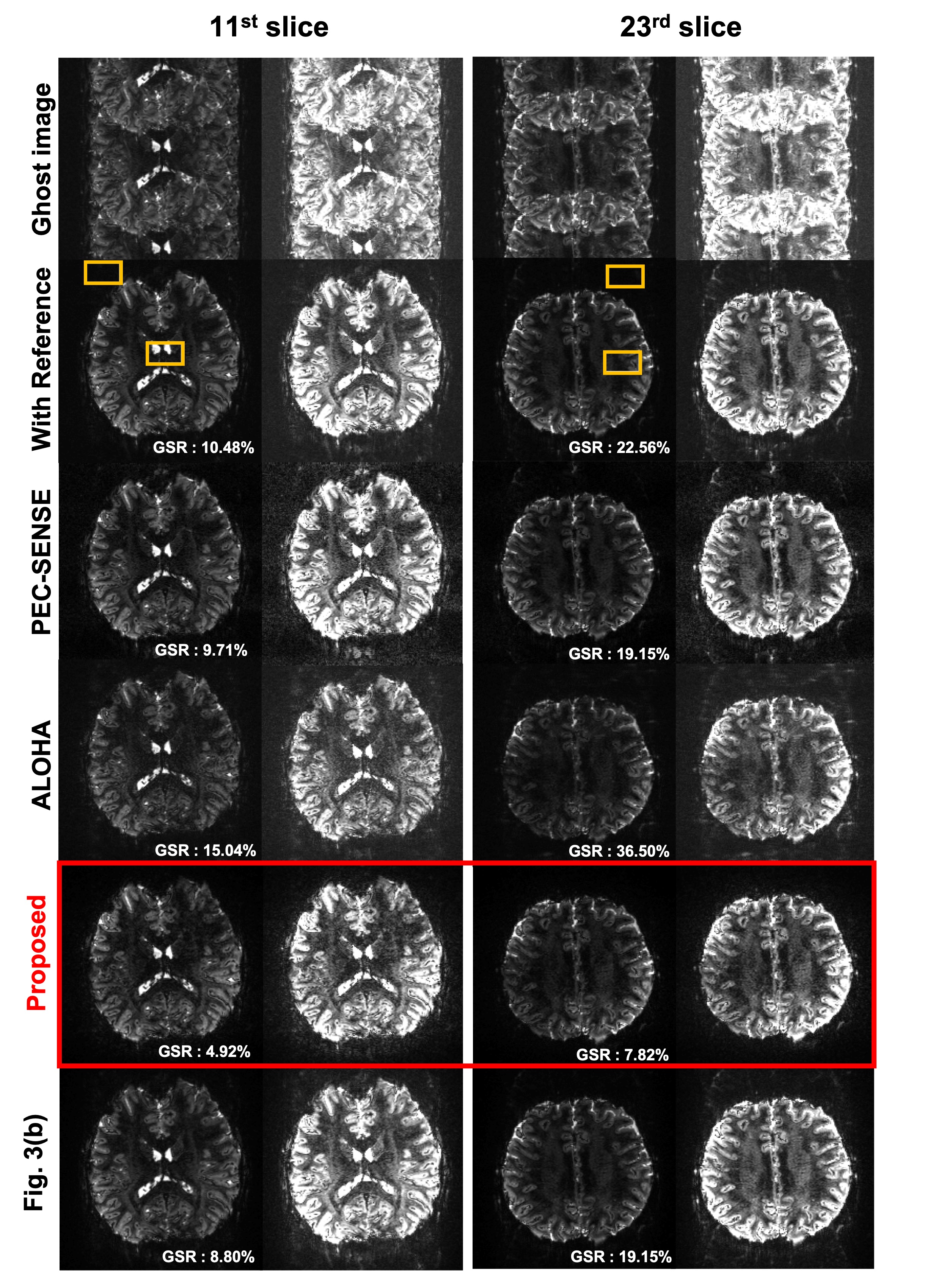}
		\caption{Ghost correction results of 7T GRE-EPI in vivo data by various methods. The last two rows correspond to the reconstruction by the methods  using Fig.~\ref{fig:twomethod}(a) and (b), respectively.
			For each slice, the 2nd column shows the intensity rescaled image of 5 times. The remaining ghost artifact is shown in other method results except the proposed method. ROI for calculating signal and ghost values are depicted by yellow rectangular.}
		\label{fig:result_7T}
	\end{figure}

	It is known that DPG  \cite{Hoge2016dual} can correct ghost artifact not only for full-sampled EPI data, but also for down-sampled data, by estimating appropriate $k$-space interpolation kernels. To implement DPG, the calibration scans should be acquired in two different directions. On the other hand, the reconstruction results in Fig.~\ref{fig:result_7T}  using the implementation of Fig.~\ref{fig:twomethod}(a) confirmed that the proposed neural network can remove ghost artifact and learn the interpolation kernels from accelerated acquisition at the same time.
	Specifically, we even found that the neural network in Fig.~\ref{fig:twomethod}(b) with explicit GRAPPA kernel estimation using calibration data is more difficult to train compared to Fig.~\ref{fig:twomethod}(a).  For example, the neural network in  Fig.~\ref{fig:twomethod}(b) required a pre-training step  with reference-based correction data as label, before  ALOHA-based correction images is used for fine tuning as label. On the other hand, the network in Fig.~\ref{fig:twomethod}(a) was successfully trained with only ALOHA-based correction images. Moreover, in  Fig.~\ref{fig:result_7T}, the reconstruction results by Fig.~\ref{fig:twomethod}(b) shows a slightly remaining ghost artifact, which results in significantly larger GSR value than the method in Fig.~\ref{fig:twomethod}(a).
	This again confirms that our neural network Fig.~\ref{fig:twomethod}(a)  can accurately estimate  $k$-space interpolation kernel estimation for both phase error correction and acceleration without any dual polarity calibration data. Therefore, the proposed method can be considered as a learning-based extension of DPG without calibration data.
	
	The prosed method could be extended for  ghost correction problem in simultaneous multislice (SMS) imaging.
	SMS is another complicated ghost correction problem, since the phase disparity between positive and negative echo depends on each slice, which
	results in ghost artifacts on the collapsed slices image. In \cite{lyu2018robust,liu2019pecgrappa}, the authors calculated the phase difference map from ghost free positive- and negative-echo images which are reconstructed by virtual coil simultaneous autocalibration and k-space estimation (VC-SAKE). Then,  SMS image is reconstructed using PEC-SENSE\cite{lyu2018robust}, and PEC-GRAPPA\cite{liu2019pecgrappa} with the estimated phase error map.
	Since our proposed network could learn both phase correction factor and interpolation kernel at the same time in Fig.~\ref{fig:twomethod}(a),  the proposed neural network may
	be useful   for SMS EPI reconstruction with Nyquist ghost correction problem.

	\section{Conclusion}
	
	In this study, we proposed a new ghost artifact correction method using $k$-space deep learning. The proposed $k$-space  deep learning  was derived based on the recent mathematical discovery that a deep convolutional neural network is related to a Hankel matrix decomposition. The ghost correction results using 3T and 7T EPI data confirmed that the proposed method  outperformed the existing approaches. Moreover, the proposed method were very robust and successfully removed the ghost artifacts even when ALOHA-based approach failed.  Furthermore, even from the accelerated EPI data, the proposed method correctly	estimated the $k$-space interpolation kernel and corrected the phase errors.

	\section*{Acknowledgment}
	We thank Jooyun Kim and Sungho Tak in Korea Basic Science Institute for assistance with acquiring MR data. This work is supported by National Research Foundation of Korea, Grant number NRF-2016R1A2B3008104.

	\pagebreak
	
	\section*{Supporting Information}
	Additional Supporting Information may be found in the online version of this article.
	\begin{figure}[h]
		\centering
		\includegraphics[width=10cm]{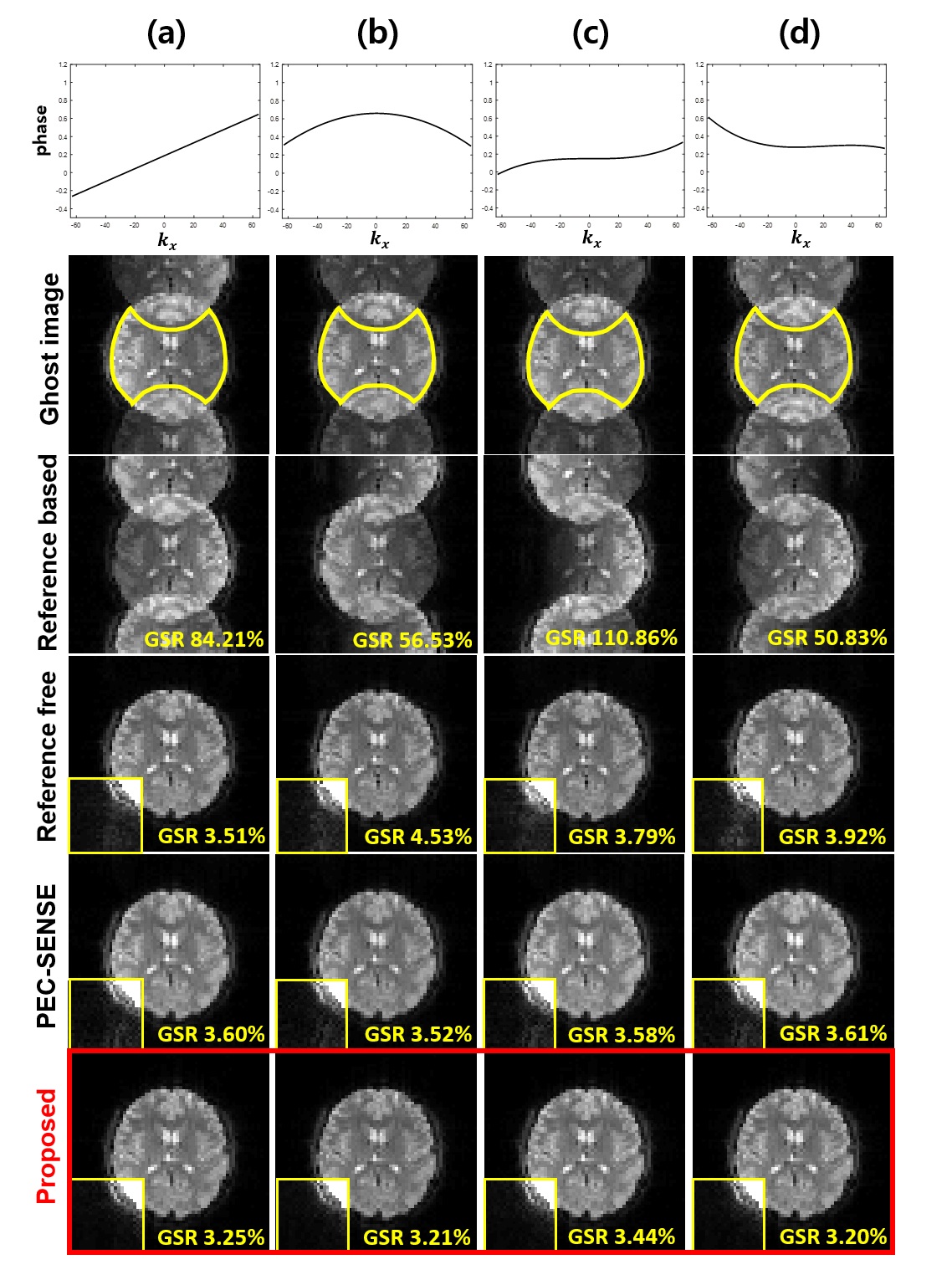}

	\end{figure}

	\textbf{FIGURE S1}   Ghost correction results for simulated non-linear phase error data. Those data are based on 3T data and 
	the phase errors are modeled as (a) a linear with constant error, (b) a quadratic with constant error, (c) a cubic with constant error, and (d) the combination of cubic, quadratic, and constant error terms. Yellow boxes show the intensity rescaled images of 5 times.  To get the GSR values, signal value is calculated inside the yellow line, and ghost value is calculated outside the brain. Here, the proposed method shows the most robust performance.
	\clearpage

	\section*{List of figures}
	
	\clearpage
	
\end{document}